\pdfoutput=1
\documentclass[11pt]{article}

\usepackage[final]{naacl2021}

\usepackage{times}
\usepackage{latexsym}

\usepackage[T1]{fontenc}
\usepackage[utf8]{inputenc}

\usepackage{microtype}
\usepackage{tikz}           % aded by author
\usetikzlibrary{shapes.geometric, arrows}
\usepackage{pgfplotstable}  % added by author for LATEX plots
\usepackage{glossaries}     % added by author for acronym.tex
\newacronym{ann}{ANN}{Artificial Neural Network}
\newacronym{ar}{'ar'}{Arabic}
\newacronym{bert}{BERT}{Bidirectional Encoder Representations from Transformers}
\newacronym{br}{'br'}{Breton}
\newacronym{de}{'de'}{German}
\newacronym{en}{'en'}{English}
\newacronym{eo}{'eo'}{Esperanto}
\newacronym{es}{'es'}{Spanish}
\newacronym{gpt}{GPT}{Generative Pre-Training}
\newacronym{fi}{'fi'}{Finnish}
\newacronym{fr}{'fr'}{French}
\newacronym{fro}{'fro'}{French Ortofasil}
\newacronym{ipa}{IPA}{International Phonetic Alphabet}
\newacronym{it}{'it'}{Italian}
\newacronym{ko}{'ko'}{Korean}
\newacronym{lstm}{LSTM}{Long Short Term Memory}
\newacronym{nl}{'nl'}{Dutch}
\newacronym{pt}{'pt'}{Portuguese}
\newacronym{nlp}{NLP}{Natural Language Processing}
\newacronym{nmt}{NMT}{Neural Machine Translation}
\newacronym{oteann}{OTEANN}{Orthographic Transparency Estimation with an ANN}
\newacronym{ru}{'ru'}{Russian}
\newacronym{sh}{'sh'}{Serbo-Croatian}
\newacronym{tr}{'tr'}{Turkish}
\newacronym{zh}{'zh'}{Chinese}
\newacronym{eno}{'eno'}{Entirely Opaque}
\newacronym{ent}{'ent'}{Entirely Transparent}
\newacronym{seq2seq}{seq2seq}{Sequence-to-sequence}       % added by author
\usepackage{floatrow}       % added by author
\usepackage{tipa}           % added by author for phonetic letter
\title{OTEANN: Estimating the Transparency of Orthographies with an Artificial Neural Network}

\author{Xavier Marjou \\
  Lannion, Brittany, France \\
  \texttt{xavier.marjou@gmail.com}}

\begin{document}
\maketitle
\begin{abstract}
To transcribe spoken language to written medium, most alphabets enable an unambiguous sound-to-letter rule. However, some writing systems have distanced themselves from this simple concept and little work exists in \gls{nlp} on measuring such distance. In this study, we use an \gls{ann} model to evaluate the transparency between written words and their pronunciation, hence its name \gls{oteann}. Based on datasets derived from Wikimedia dictionaries, we trained and tested this model to score the percentage of correct  predictions in phoneme-to-grapheme and grapheme-to-phoneme translation tasks. The scores obtained on 17 orthographies were in line with the estimations of other studies. Interestingly, the model also provided insight into typical mistakes made by learners who only consider the phonemic rule in reading and writing.
\end{abstract}

\section{Introduction}
An alphabet is a standard set of letters that represent the basic significant sounds of the spoken language it is used to write. When a spelling system (also referred as \emph{orthography}) systematically uses a one-to-one correspondence between its sounds and its letters, the encoding of a sound (also referred as \emph{phoneme}) into a letter (also referred as \emph{grapheme}) leads to a single possibility; similarly the decoding of a letter into a sound leads to a single possibility as well. Such orthography is thus \emph{transparent} with regards to phonemes with the advantage of offering no ambiguity when writing or reading the letters of a word, as illustrated in Figure \ref{fig:unambiguous-spelling-system}.

\begin{figure*}
    \centering
    
\begin{tikzpicture}

\definecolor{myolive}{rgb}{0.8,0.725,0.454}
\definecolor{myviolet}{rgb}{0.505,0.447, 0.701}
\tikzstyle{sbox} = [rectangle, rounded corners, minimum width=3cm, minimum height=3.5cm,text centered, draw=black, fill=myolive!30]
\tikzstyle{lbox} = [rectangle, rounded corners, minimum width=3cm, minimum height=3.5cm,text centered, draw=black, fill=myviolet!30]
\tikzstyle{label} = [minimum width=0.7cm, minimum height=0.7cm,text centered]
\tikzstyle{symbol} = [minimum width=0.7cm, minimum height=0.7cm,text centered]
\tikzstyle{arrow} = [dashed,->,>=stealth]

\node (box1) [sbox, yshift=1.5cm] {};
\node (box2) [lbox, right of=box1, xshift=2.5cm] {};
\node (s1) [symbol, yshift=1.5cm] {/t/};
\node (l1) [symbol,right of=s1, xshift=2.5cm, yshift=1.2cm] {<t>};
\node (l2) [symbol,right of=s1, xshift=2.5cm, yshift=0.8cm] {<t>+<e>};
\node (l3) [symbol,right of=s1, xshift=2.5cm, yshift=0.4cm] {<t>+<t>};
\node (l4) [symbol,right of=s1, xshift=2.5cm, yshift=0cm] {...};
\node (l5) [symbol,right of=s1, xshift=2.5cm, yshift=-0.4cm] {<t>+<h>};
\node (l6) [symbol,right of=s1, xshift=2.5cm, yshift=-0.8cm] {<t>+<e>+<s>};
\node (l7) [symbol,right of=s1, xshift=2.5cm, yshift=-1.2cm] {<t>+<e>+<n>+<t>};
\draw [arrow] (s1) -- (l1);
\draw [arrow] (s1) -- (l2);
\draw [arrow] (s1) -- (l3);
\draw [arrow] (s1) -- (l4);
\draw [arrow] (s1) -- (l5);
\draw [arrow] (s1) -- (l6);
\draw [arrow] (s1) -- (l7);

\node (box3) [lbox, right of=box2, xshift=3cm] {};
\node (box4) [sbox, right of=box3, xshift=2.5cm] {};
\node (ll1) [symbol, xshift= 7.5cm, yshift=1.5cm] {<t>};
\node (ss1) [symbol, right of=ll1, xshift=2.5cm, yshift=0.4cm] {/t/};
\node (ss2) [symbol, right of=ll1, xshift=2.5cm, yshift=0cm] {/s/};
\node (ss3) [symbol, right of=ll1, xshift=2.5cm, yshift=-0.4cm] {/s/+/y/};
\draw [arrow] (ll1) -- (ss1);
\draw [arrow] (ll1) -- (ss2);
\draw [arrow] (ll1) -- (ss3);

\end{tikzpicture}
\caption{Example of ambiguous correspondence during writing and reading tasks in French. The /t/ phoneme can correspond to multiple graphemes, depending on the nature of the word and also depending on the nature of neighboring words in the sentence or even in a previous sentence. Similarly, the <t> grapheme can correspond to multiple phonemes.}
\label{fig:ambiguous-spelling-system}
\end{figure*}
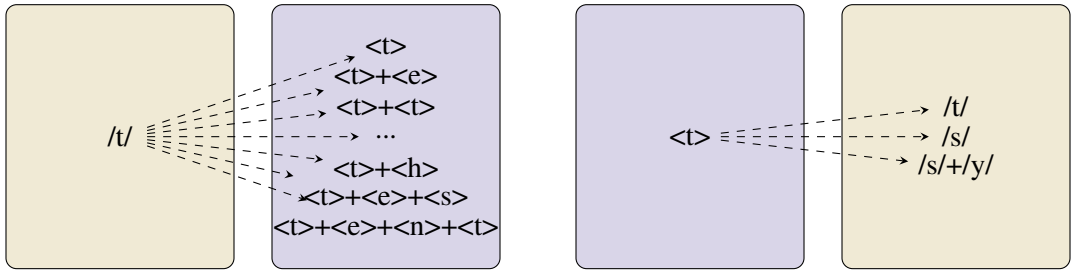

In real life, no existing orthography is fully transparent phonemically. One reason is that a word spoken alone is sometimes different from a word spoken in a sentence. An even more consequential reason is that some orthographies like English\footnote{\url{https://en.wikipedia.org/wiki /English_orthography\#Spelling_patterns}} and French\footnote{\url{https://fr.wiktionary.org/wiki/Annexe:Prononciation/français}} have incorporated deeper depth rules that have moved them away from a transparent orthography \cite{seymour2003foundation}; this has created ambiguities when trying to write or read phonemically, as illustrated in Figure \ref{fig:ambiguous-spelling-system}. 

\suppressfloats % added to avoid picture in the (top of the) first page

\begin{figure*}
\centering

\begin{tikzpicture}

\definecolor{myolive}{rgb}{0.8,0.725,0.454}
\definecolor{myviolet}{rgb}{0.505,0.447, 0.701}
\tikzstyle{sbox} = [rectangle, rounded corners, minimum width=3cm, minimum height=1cm,text centered, draw=black, fill=myolive!30]
\tikzstyle{lbox} = [rectangle, rounded corners, minimum width=3cm, minimum height=1cm,text centered, draw=black, fill=myviolet!30]
\tikzstyle{label} = [minimum width=0.7cm, minimum height=0.7cm,text centered]
\tikzstyle{symbol} = [minimum width=0.7cm, minimum height=0.7cm,text centered]
\tikzstyle{arrow} = [thick,->,>=stealth]

\node (box1) [sbox, yshift=1.5cm] {};
\node (box2) [lbox, right of=box1, xshift=2.5cm] {};
\node (s1) [symbol, yshift=1.5cm] {/t/};
\node (l1) [symbol,right of=s1, xshift=2.5cm, yshift=0cm] {<t>};
\draw [arrow] (s1) -- (l1);

\node (box3) [lbox, right of=box2, xshift=3cm] {};
\node (box4) [sbox, right of=box3, xshift=2.5cm] {};
\node (ll1) [symbol, xshift= 7.5cm, yshift=1.5cm] {<t>};
\node (ss1) [symbol, right of=ll1, xshift=2.5cm, yshift=0cm] {/t/};
\draw [arrow] (ll1) -- (ss1);

\end{tikzpicture}
\caption{Example of unambiguous correspondence during writing and reading tasks in Esperanto.}
\label{fig:unambiguous-spelling-system}
\end{figure*}
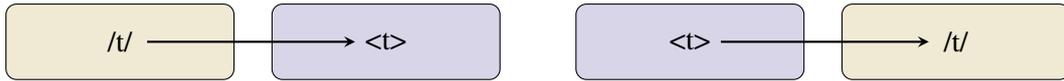

Many studies have discussed the degree of transparency of orthographies \cite{borleffs2017measuring}. These studies are mainly motivated by the estimation of the ease of reading and writing when learning a new language \cite{defior_martos_cary_2002}. Finnish, Korean, Serbo-Croatian and Turkish orthographies are often referred as highly \emph{transparent} \cite{aro2004learning} \cite{wang2009rule}, \cite{turvey1984serbo}, \cite{oney1997beginning}, whereas English and French orthographies are referred as \emph{opaque} \cite{Bosch_Analysingorthographic}. However, little work exists in NLP about measuring the level of transparency of an orthography. One noticeable exception is the work of \citet{Bosch_Analysingorthographic} who have created grapheme-to-phoneme scores and tested them on three orthographies (Dutch, English and French).

This study extends such work with a method called \gls{oteann}, which models a word-based \emph{phoneme-to-grapheme} task and a word-based \emph{grapheme-to-phoneme} task using an \gls{ann}. For the sake of simplicity, the former task is called a \emph{writing} task while the latter task is called a \emph{reading} task. The goal is not to build a perfect spelling translator or a spell checker. Instead the goal is to build a translator which can indicate a degree of phonemic transparency and thus make it possible to rank orthographies according to this criterion. 

Interestingly, recent years have seen tremendous progress regarding \gls{nlp} with \gls{ann}s \cite{otter2018survey}. \citet{DBLP:journals/corr/SutskeverVL14} proposed an \gls{ann} called a \gls{seq2seq} model that has proven to be very successful on language translation tasks. More recently, \gls{ann}s based on as \emph{attention} \cite{bahdanau2014neural}, \cite{vaswani2017attention} and \emph{transformers} like \gls{bert}, \cite{devlin2018bert} and \gls{gpt} \cite{Radford2018ImprovingLU} have again enhanced and outperformed \gls{seq2seq}s. Considering writing a word and reading a word as two translations tasks allows re-using the transformers for our work. To this purpose, we used a minimalist \gls{gpt} implementation \cite{minGPT} called \emph{minGPT}. Notice that since we don't aim at building a perfect spelling translator, we do not have to translate a sequence of words into another sequence of words; our model only requires translating a spoken word into a spelled word (\emph{writing task}) and a spelled word into a spoken word (\emph{reading task}). In other words, our \gls{ann} operates at the character level within a sequence of characters of single words. The pronunciation and spelling of the word are both encoded as a sequence of UTF-8 characters; a pronounced word is encoded with the characters belonging to the set of phonemes of the target language, whereas a spelled word is encoded with the characters belonging to the alphabet of the target orthography. 
We directly re-used minGPT code with no modification. The only differences were the training data and the code for extracting the prediction at inference time.

We used \gls{oteann} to test seventeen orthographies in order to evaluate their degree of phonemic transparency. Sixteen of them are the official orthographies of their respective language (Arabic, Breton, Chinese, Dutch, English, Esperanto, Finnish, French, German, Italian, Korean, Portuguese, Russian, Serbo-Croatian, Spanish, and Turkish) while the seventeenth is a phonemic orthography proposed for French.

A unique multi-orthography \gls{ann} model instance was trained to learn the writing and reading tasks on all languages at the same time. In other words, we used a single dataset containing samples of all studied orthographies. The multi-orthography \gls{ann} model was then tested for each orthography and each task with new samples, which allowed calculating an average percentage of correct translations. A score of $0\%$ of correct translations represented a fully opaque orthography (no correlation between the input and the target), whereas a score close to $100\%$ represented a fully transparent orthography (full correlation between the input and the target).

Our study first confirms that orthographies like Arabic, Finnish, Korean, Serbo-Croatian and Turkish are highly transparent whereas other ones like Chinese, French and English are highly opaque. For example, when solely based on a phoneme-grapheme correspondence, we estimated the chances of correctly writing a French word at $28\%$; similarly, when solely based on a grapheme-phoneme correspondence, we estimated the chances of correctly pronouncing an English word at $31\%$. For Dutch, English and French reading tasks, our obtained ranking is in line with the one of \citet{Bosch_Analysingorthographic}. One unexpected finding is that \gls{oteann} also allows discovering certain mistakes performed by a new learner during writing and reading. 

Remarkably, our method should apply to any orthography, provided a dataset is available.

\section{Methodology} \label{methodology}

In order to evaluate a level of transparency of some orthographies two main steps were necessary: obtaining datasets and carrying out the training and testing experiments with the \gls{ann}.

\subsection{Datasets} \label{dataset}

As displayed in Table \ref{tab:dataset_features}, we needed a multi-orthography dataset with four features per sample: the orthography, the task (write or read), the input word (pronunciation or spelled word) and the output word (spelled word or pronunciation). A spelled word was represented by a sequence of graphemes whereas a pronunciation was represented by a sequence of phonemes. The characters representing phonemes are also called \gls{ipa} characters. Having a single dataset with multiple orthographies and tasks allows a single multi-orthography \gls{ann} model to learn to read and write all orthographies; otherwise, it would require one \gls{ann} model per orthography-task pair.

\begin{table}
  \begin{center}
    \begin{tabular}{c|c|c|c} 
      \textbf{Orthography} & \textbf{Task} & \textbf{Input} & \textbf{Output} \\
      \hline
      en & write & \textipa{dZ6b} & job \\
      en & read & job & \textipa{dZ6b} \\
    \end{tabular}
    \caption{Features of the multi-orthography dataset}
    \label{tab:dataset_features}
  \end{center}
\end{table}

In order to build such dataset, we first generated one sub-dataset per orthography (e.g. one 'en' sub-dataset for English), each containing the pronunciation and the spelled word (e.g. '\textipa{dZ6b}' and 'job'). 

\subsubsection{Baseline Orthographies}

We first created baselines representing a fully transparent orthography and a fully opaque orthography.

Regarding a fully transparent orthography, we created a new artificial orthography called \gls{ent} orthography. We generated its samples by using the \gls{ipa} pronunciation of real Esperanto words both as the pronunciation and as the spelled word, which resulted in a sub-dataset containing an \gls{ent} \emph{bijective} orthography.

Regarding a fully opaque orthography, we also created a new artificial orthography called \gls{eno} orthography. We generated its samples by taking the \gls{ipa} pronunciation of real Esperanto words mapping each of theirs phonemes to a random grapheme from a list of $25$ graphemes, which resulted in a sub-dataset containing an \gls{eno} orthography with no correlation between the pronunciation and the spelled word.

\subsubsection{Studied Orthographies}

A sub-dataset was created for each of the following orthographies: \gls{ar}, \gls{br}, \gls{de}, \gls{en}, \gls{eo}, \gls{es},  \gls{fi}, \gls{fr}, \gls{it}, \gls{ko}, \gls{nl}, \gls{pt}, \gls{ru}, \gls{sh}, \gls{tr} and \gls{zh}.

We incorporated the words from the corresponding Wiktionary\footnote{\url{https://wiktionary.org}} dump\footnote{\url{https://dumps.wikimedia.org/}}, with the exception of the following ones:
\begin{itemize}
    \item Words containing space characters;
    \item Words containing more than 25 characters;
    \item Words containing capital letters (except for German words);
    \item Words containing non-standard characters with regard to the orthography's alphabet.
\end{itemize}

Two orthographies required additional processing: 
\begin{itemize}
\item For German, proper nouns were discarded and the capital letter of common nouns was transformed into lower case;
\item For Korean, the syllabic blocks words were converted in a series of two or three letters (one vowel and one or two consonants) pertaining to the Korean alphabet with \emph{ko\_pron}\footnote{\url{https://pypi.org/project/ko-pron}} Python library.
\end{itemize}

Regarding pronunciation, we directly extracted the \gls{ipa} pronunciation when available in the associated Wiktionary dump, which was the case for 'br', 'de', 'en', 'es', 'fr', 'it', 'nl', 'pt' and 'sh'. The Esperanto ('eo') pronunciation came from the French Wiktionary. For the others  ('ar', 'ko', 'ru', 'fi', 'tr'), we had to derive it from the spelled word with additional software. For Russian, the Russian Wiktionary dump did not contain the \gls{ipa}. We thus used \emph{wikt2pron} \emph{ru\_pron} module\footnote{\url{https://wikt2pron.readthedocs.io/en/latest/_modules/IPA/ru_pron.html}} to obtain a pronunciation similar to the one displayed in the Russian Wiktionary web pages. For Chinese, we only selected Mandarin words in simplified Chinese and limited to one or two symbols (a.k.a. Hanzis); we then obtained their pronunciation from the \emph{CEDICT}\footnote{\url{https://github.com/msavva/transphoner/blob/master/data/ }} dataset.

Extracting the phonemic pronunciation from Wiktionary may raise concerns given than \gls{ipa} symbols can be used both for phonetic and phonemic notations and that there is no unified consistency between the different dictionaries. When processing the \gls{ipa} strings, we nonetheless took care of preserving the highest surface pronunciation as possible: most pitches were removed since they represent no useful hint during the writing task (i.e. no consequence on the spelled word) and especially since they are generally impossible to predict when translating the spelled word into a pronunciation during the reading task. Nevertheless the \textipa{/:/} pitch was noticed as indispensable for some orthographies, for instance for predicting double vowels in the spelling of Finnish words or the \textit{alif} letter in Arabic. Regarding the \textipa{/"/} pitch, it can slightly influence Spanish translation scores: it can lead to a better writing score as it can be a hint for predicting accented letters, but it can also lead to a lower reading score.

Another interesting orthography was a proposal of an alternative orthography for French called \gls{fro}\footnote{\url{https://fonétik.fr/v0/faq-en.html\#mapping-table}}, which seeks to be phonemically transparent. Although not fully bijective (e.g. both /\textipa{o}/ and /\textipa{O}/ map to <o> letter), it indeed seems highly transparent. We therefore used it to generate a sub-dataset for the \gls{fro} orthography. 

It is debatable whether Chinese should be included in this study given the term \emph{alphabet} is usually reserved for largely phonographic systems that have a small number of elements. We decided to include it because our \gls{ann} model allowed for \emph{alphabets} with thousands of graphemes. 

Table \ref{tab:datasets_summary} summarizes the sub-datasets obtained. 

%Some Wiktionaries like the French Wiktionary also indexes a few words that are often incorrectly spelled\footnote{\url{https://fr.wiktionary.org/wiki/Catégorie:Termes_non_standards_en_français}}. This is no problem since they are statistically insignificant (e.g. 189 such words out of 1,214,262 for \gls{fr}).

\begin{table*}
  \begin{center}
    \begin{tabular}{c|c|c|c|c|c} % <-- Alignments: 1st column left, 2nd middle and 3rd right, with vertical lines in between
      \textbf{Orthography} & \textbf{Samples} & \textbf{Phonemes} & \textbf{Graphemes} & \textbf{Nb. of Phonemes} & \textbf{Nb of Graphemes} \\
      \hline
ar & 12,057 & 32 & 47 & 8.0 ± 2.0 & 8.9 ± 2.3 \\
br & 17,343 & 45 & 29 & 6.6 ± 1.9 & 7.5 ± 2.2 \\
de & 529,740 & 41 & 30 & 10.2 ± 3.1 & 11.5 ± 3.4 \\
en & 42,206 & 42 & 29 & 7.3 ± 2.7 & 7.6 ± 2.6 \\
eo & 26,845 & 25 & 28 & 8.8 ± 2.6 & 8.6 ± 2.5 \\
es & 40,824 & 34 & 33 & 8.1 ± 2.7 & 8.7 ± 2.6 \\
fi & 105,352 & 28 & 27 & 10.4 ± 3.5 & 10.4 ± 3.5 \\
fr & 1,214,248 & 35 & 41 & 9.0 ± 2.7 & 11.2 ± 2.9 \\
fro & 1,214,262 & 35 & 32 & 9.0 ± 2.7 & 8.6 ± 2.6 \\
it & 26,798 & 34 & 32 & 9.1 ± 2.8 & 9.1 ± 2.6 \\
ko & 64,669 & 41 & 67 & 10.6 ± 4.0 & 8.3 ± 3.0 \\
nl & 13,340 & 45 & 28 & 7.8 ± 3.1 & 8.6 ± 3.4 \\
pt & 12,190 & 37 & 38 & 7.7 ± 2.3 & 7.9 ± 2.3 \\
ru & 304,514 & 30 & 33 & 10.5 ± 3.1 & 10.7 ± 3.1 \\
sh & 98,575 & 27 & 27 & 9.1 ± 2.8 & 8.9 ± 2.7 \\
tr & 117,841 & 36 & 31 & 10.3 ± 3.7 & 10.1 ± 3.6 \\
zh & 27,688 & 32 & 4813 & 9.9 ± 2.2 & 1.8 ± 0.3 \\
eno & 26,845 & 25 & 25 & 8.8 ± 2.6 & 8.8 ± 2.6 \\
ent & 26,845 & 25 & 25 & 8.8 ± 2.6 & 8.8 ± 2.6 \\
    \end{tabular}
  \end{center}
  \caption{Summary of the sub-datasets. For each sub-dataset, a line indicates the number of samples available, the number of different phoneme UTF-8 characters, the number of different grapheme UTF-8 characters, the mean number of phonemes in words, and the mean number of graphemes in words.}
  \label{tab:datasets_summary}
\end{table*}

\subsubsection{Training and test datasets}

$11,000$ samples were randomly selected in each of the $17$ sub-datasets. Each sample from a sub-dataset produced two samples in the multi-orthography dataset: one sample for \emph{write} task and one sample for the \emph{read} task, as illustrated in Table \ref{tab:dataset_features}. This  multi-orthography dataset was subsequently divided into a training dataset ($10,000$ * $17$ * $2$ samples) and a test dataset ($1,000$ * $17$ * $2$ samples). 

\subsubsection{ANN architecture}

We used minGTP \cite{minGPT} which runs on PyTorch\footnote{\url{https://pytorch.org/}}. Regarding the hyper-parameters, we configured a block size of 63 characters, 4 layers, 4 heads and 336 embedding tokens, which resulted in an \gls{ann} of $9,589,536$ trainable parameters and an episode training time of 2 hours and 10 minutes on a 4 GPU node. No effort was spent to shrink or prune the \gls{ann}, so its size could still be optimized. The data and code are available on Github \footnote{\url{https://github.com/marxav/oteann4}}.

\subsubsection{Performance metric}

We used a simple score in order to assess the performance of the \gls{ann} prediction during the testing step. When all the predicted characters were equal to those of the true target, a prediction was considered successful, hence allowing to score the percentage of successful predictions performed for each orthography-task pair.

\subsubsection{Training and testing}

We specified an episode as:
\begin{itemize}

\item{
\textbf{Generating the training and test datasets.}
At the end of this step, each character present in these datasets was provisioned in the inventory of the \gls{ann} instance.
} 

\item{
\textbf{Training the \gls{ann} model}. The full training dataset was processed to be used as text blocks containing the concatenation of the four features (\emph{orthography}, \emph{task}, \emph{input} and \emph{output}) separated by a comma. Therefore, a single instance of the model was used to learn to write and read all 17 orthographies in one training.}

\item{
\textbf{Testing the \gls{ann} model for each orthography-task pair}. For each orthography-task pair, $1,000$ new samples were tested. Each sample was fed into the model with the concatenation of the three first features (\emph{orthography}, \emph{task} and \emph{input}) separated by a comma. The model had to predict a value equal to the \emph{output} feature, which was the target to be found.}

\end{itemize}

We performed 11 episodes to measure the mean and standard deviation of each orthography-task pair and thus assess the consistency of our results.

Future work may use more test samples to gain a statistical insight on the different types of errors depending on the orthography at hand.

\section{Results}  \label{results}

First, regarding the results of the two baseline orthographies, the \gls{eno} opaque orthography obtained a score of $0\%$ in both writing and reading, which was in line with the expectations given that there was no correlation between its phonemes and its graphemes; on the other hand, the \gls{ent} transparent orthography scored above $99.6\%$ on the writing and reading tasks, which indicated a high level of correlation between its phonemes and its graphemes. We thus considered our \gls{ann} model satisfactory for our objective of comparing the performance of different orthographies.

\begin{figure}[h]
\centering
\captionsetup{justification=centering}
\ffigbox{%
\includegraphics[width=7.5cm]{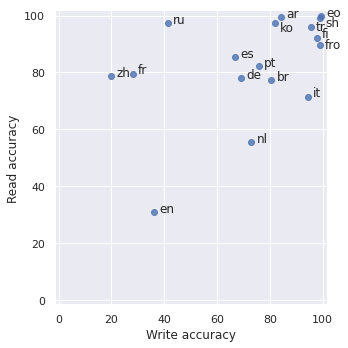}%
}{%
  \caption{Scatterplot of the mean scores. \newline (OTEANN trained with $10,000$ samples)}%
  \label{fig:figure_results}
}
\end{figure}

Figure \ref{fig:figure_results}  and Table \ref{tab:tabular_results} present our main results. They are significantly different between writing and reading since these tasks are generally not symmetrical. Two features are likely to influence the symmetry, and therefore the efficiency of each task. As recalled by Figure \ref{fig:ambiguous-spelling-system}, the most important feature would undoubtedly be the number of possible phoneme-to-grapheme and grapheme-to-phoneme ambiguities per tested orthography. Unfortunately we did not possess such data. Another impacting feature may be the number of possible values (graphemes or phonemes) for a given target character. The higher the number of values, the harder the prediction should be for the \gls{ann}. Future work should investigate the relative importance of these features on the \gls{oteann} performances.

\begin{table}[h]
\centering
\captionsetup{justification=centering}
\begin{tabular}{c c c}
\hline
\textbf{Orthography} & \textbf{Write} & \textbf{Read}\\
\hline\hline
ent & 99.6 ± 0.3 & 99.8 ± 0.1 \\\hline
eno &  0.0 ± 0.0 &  0.0 ± 0.0 \\\hline
\hline
ar  & 84.3 ± 0.8 & 99.4 ± 0.3 \\\hline
br  & 80.6 ± 0.6 & 77.2 ± 1.6 \\\hline
de  & 69.1 ± 1.0 & 78.0 ± 1.5 \\\hline
en  & 36.1 ± 1.5 & 31.1 ± 1.3 \\\hline
eo  & 99.3 ± 0.2 & 99.7 ± 0.1 \\\hline
es  & 66.9 ± 2.0 & 85.3 ± 1.3 \\\hline
fi  & 97.7 ± 0.3 & 92.3 ± 0.8 \\\hline
fr  & 28.0 ± 1.4 & 79.6 ± 1.7 \\\hline
fro & 99.0 ± 0.3 & 89.7 ± 1.1 \\\hline
it  & 94.5 ± 0.8 & 71.6 ± 0.9 \\\hline
ko  & 81.9 ± 1.0 & 97.5 ± 0.5 \\\hline
nl  & 72.9 ± 1.7 & 55.7 ± 2.2 \\\hline
pt  & 75.8 ± 1.0 & 82.4 ± 0.9 \\\hline
ru  & 41.3 ± 1.6 & 97.2 ± 0.5 \\\hline
sh  & 99.2 ± 0.3 & 99.3 ± 0.3 \\\hline
tr  & 95.4 ± 0.7 & 95.9 ± 0.6 \\\hline
zh  & 19.9 ± 1.4 & 78.7 ± 0.9 \\\hline
\end{tabular}
{%
  \caption{Phonemic transparency scores. \newline (OTEANN trained with $10,000$ samples)}%
  \label{tab:tabular_results}
}

\end{table}

Comparing \gls{oteann}'s reading results with those of \citet{Bosch_Analysingorthographic}, \gls{oteann} first seems to naturally assimilate the grapheme complexity (e.g. for French, it successfully learnt that "cadeau" should be pronounced \textipa{/kado/}). Regarding grapheme-to-phoneme complexity (\emph{G-P complexity}), they ranked English (\emph{G-P complexity}=$90\%$) more complex than Dutch (\emph{G-P complexity}=$25\%$) which, in turn, was more complex than French (\emph{G-P complexity}=$15\%$). \gls{oteann} results preserved the same ranking with transparency scores of $31\%$, $57\%$ and $79s\%$ for English, Dutch and French. Admittedly, \gls{oteann}'s scores were different in terms of scale but \gls{oteann} had to deal with more orthographies as well as with the writing task.

Figure \ref{fig:figure_results} also allows categorizing the studied orthographies with respect to their degree of transparency:

\begin{itemize}

\item{\textbf{Esperanto}}: With scores above $99.3\%$, Esperanto orthography is nearly as transparent as the \gls{ent} baseline. The most common error occurred on a doubled letter in the input, which was incorrectly translated to a single letter.

\item{\textbf{Arabic, Finnish, Korean, Serbo-Croatian and Turkish}}: Their scores above $80\%$ both in writing and reading confirmed that their orthography is highly transparent as indicated in \cite{aro2004learning}, \cite{wang2009rule}  and \cite{oney1997beginning}. The Arabic score is high on in the read direction, which is likely due to the use of diacritics in the dataset; without them, the score would undoubtedly be lower. Regarding Korean, its orthography became a little less transparent during the twentieth century; its high scores suggest that further work should check the dataset and evaluate new scores.  

\item{\textbf{Breton, German, Italian, Portuguese and Spanish}}: With all their scores above $65\%$  their orthography was also measured as fairly transparent. For Spanish, the detailed results showed that the most common failure during writing occurs with accents: the \gls{ann} had great difficulty predicting whether a vowel should contain an accent or not. For Italian, typical errors observed in the results were the prediction of \textipa{/E/} instead of a \textipa{/e/} and \textipa{/O/} instead of a \textipa{/o/}, which were harder to discriminate. Future work may revise the scoring formula to reduce the cost of some of these errors in the performance calculation.

\item{\textbf{Dutch}}: The Dutch reading score  ($56\%$) is low but might be slightly enhanced given a possible lack of consistency regarding the phonemes used in the Dutch sub-dataset.

\item{\textbf{Russian}}: The Russian writing score ($41\%$) may seem low. However, Russian has strong stress-related vowel reduction, which makes it hard to know how to write a word without knowing the morphemes involved. Nevertheless, future work should either study their sub-dataset more in depth or use a different data source like {wikipron}\footnote{\url{https://pypi.org/project/wikipron/}} to possibly improve its scores. 

\item\textbf{{Chinese}}: The results indicated a low writing score ($20\%$), which is not surprising given than some phonemes can have multiple corresponding graphemes and that there are thousands of graphemes (Hanzis) to be learnt. However, it turns out that its reading score is much higher ($79\%$). 

\item\textbf{{French}}: With a low writing score ($28\%$), the results showed that the chances of correctly writing a French word on the sole basis of its pronunciation were rare, as anticipated given the high number of phoneme-to-grapheme possibilities. Without being able to access a broader context than the word itself, the \gls{ann} was not able to reliably predict how to write a French word. With a much higher reading score ($80\%$), the \gls{ann} obtained good reading results.  As a comparison, for the same language, the alternative \gls{fro} orthography obtained excellent writing score ($99\%$) and reading score ($90\%$). Recall that the difference between its two scores is due to the fact that the \gls{fro} orthography is not bijective. For instance, in the reading direction, the <o> letter can be translated into \textipa{/o/} or \textipa{/O/}). 

\item{\textbf{English}}: With a low writing score ($36\%$) and a low reading score ($31\%$), the results showed that English orthography is also highly opaque, which is consistent with most studies. As a reminder, a phonemic reading of an English word often does not work because of its high number of grapheme-to-phoneme  possibilities. For instance the grapheme <u> can either correspond to \textipa{/2/} (as in "hug"), to \textipa{/ju:/} (as in "huge"), to \textipa{/3:r/} (as in "cur") or \textipa{/jU@:/} as in "cure". As for Russian, additional work should be dedicated to check the English sub-dataset and possibly enhance it if necessary, which could improve \gls{en} scores by a few percent.

\end{itemize}

Observing the detailed result of each prediction also made it possible to study the phonemic correspondences learned or not learned by the \gls{oteann} model. 
\begin{itemize}

\item{For task-orthographies with a high transparency score, the model successfully predicted most pronunciations or spellings even when the correspondences involved more than one letter. For instance, \gls{oteann} predicted that the Italian word "cerchia" should be pronounced \textipa{/\textteshlig{}erkja/}, hence showing that the model had successfully learned that <c>, when followed by <e>, should be pronounced as \textipa{/\textteshlig{}/} and also that <c>, when followed by <h>, should be pronounced as \textipa{/k/}.}

\item{For task-orthographies with a low transparency score, the model generally failed on letters involved in ambiguous correspondences (recall Figure \ref{fig:ambiguous-spelling-system}). For instance, it incorrectly predicted that the pronunciation of the English word "level" was "\textipa{liv9l}" instead of "\textipa{lEv9l}", which might be a bad generalization from words like "lever" learned at training time. \gls{oteann} also incorrectly predicted that the spelling of the French word \textipa{/ale/} was "allez" when the expected target was "aller" (another French homophone); this type of error is inevitable since the \gls{oteann} model intentionally use single word input samples and therefore cannot rely on neighboring words as additional context to discriminate between homophones with different spelling.}

\item{Surprisingly, the model also predicted spellings that do not exist but who could have existed, in the same vein as \emph{ThisWordDoesNotExist.com}\footnote{\url{https://www.thisworddoesnotexist.com}}. For instance, \gls{oteann} predicted that the spelling of the French word \textipa{/swaKe/}" was "soirer", which does not exist but looks like a French infinitive verb that would mean "to celebrate at a party".
}
\end{itemize}

In addition, the results in Table \ref{tab:tabular_results}  also showed that the ANN has less than a $30\%$ chance of correctly writing a word in French or Chinese after training on 10000 samples while Figure \ref{fig:figure_training_samples_results} shows that the same ANN has more than a $85\%$ chance of correctly writing a word in Finnish, Italian, Serbo-Croatian or Turkish after training only on 1000 samples. Such a discrepancy highlights the enormous additional cost in terms of time and energy for learning a non-transparent orthography. 

\section{Discussion and Conclusion}

Among the tested orthographies, some shared the grapheme inventory. Given that they are all trained together, there might be an impact on performance. Although some of our preliminary experiments with a single \gls{ann} instance per orthography did not seem to lead to significant differences, it could be interesting to formally compare both approaches.

The accuracy metric we used is all or nothing. Additional work could also study alternative accuracy metrics and compare their results on the different orthographies. 

Although Wiktionary data may be inconsistent in quality and therefore positively or negatively impact the measured metric, the results obtained for Dutch, English and French orthographies reasonably extended those of \citet{Bosch_Analysingorthographic} while the other results reflected the perception of several other studies. Consequently, our \gls{oteann} model showed that an \gls{ann} can convincingly estimate a level of phonemic transparency for multiple orthographies both for the phoneme-to-grapheme and grapheme-to-phoneme directions.

This method should be easily applicable to other orthographies beyond those tested in this study.  However, since the superfluous \gls{ipa} symbols slightly influence the score results, future work should closely examine and discuss the phonemes to use depending on the orthography to be tested. 

As \gls{oteann} also points out some possible grapheme or phoneme errors when writing or reading phonemically, it could also be used to detect possible errors in the dictionaries of transparent orthographies; it could also be used to evaluate proposals for improving opaque orthographies.

Finally, it would be beneficial to investigate if our \gls{ann} and its artificial neural units somehow imitate the way a beginner learns to write and read a language. If so, it might suggest that a transparent orthography would be easier and faster to learn than an opaque orthography.

% Entries for the entire Anthology, followed by custom entries
\bibliography{anthology,custom}
\bibliographystyle{acl_natbib}

\appendix
\section{Additional Experiments and Results}
\label{sec:appendix}

\begin{figure}
\centering
\captionsetup{justification=centering}
\includegraphics[width=7cm]{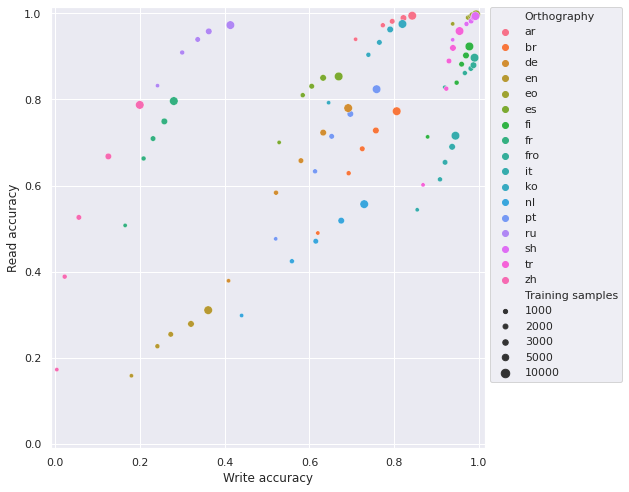}%
\caption{Scores according to the number of training samples}%
\label{fig:figure_training_samples_results}
\end{figure}

In addition to testing our \gls{oteann} model trained on $10,000$ samples, we also tested the same \gls{oteann} model but trained with fewer samples ($1,000$, $2,000$, $3,000$, and $5,000$), each time following the methodology described in section \ref{methodology}. We then aggregated the results to summarize them in Figure \ref{fig:figure_training_samples_results}, which shows the learning curve of the studied orthographies as a function of the number of training samples.

\begin{figure}
\centering
\captionsetup{justification=centering}
\ffigbox{%
\includegraphics[width=7cm]{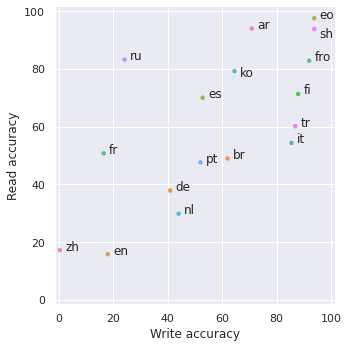}%
}{%
  \caption{Scores with $1,000$ training samples.}%
  \label{fig:figure_results_1000}
}
\end{figure}

\begin{figure}
\centering
\captionsetup{justification=centering}
\ffigbox{%
\includegraphics[width=7cm]{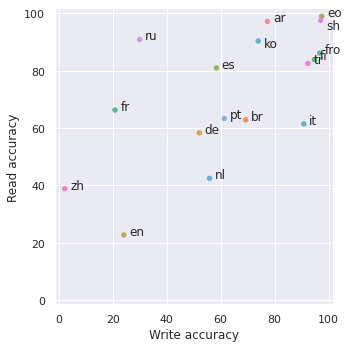}%
}{%
  \caption{Scores with $2,000$ training samples.}%
  \label{fig:figure_results_2000}
}
\end{figure}

\begin{figure}
\centering
\captionsetup{justification=centering}
\ffigbox{%
\includegraphics[width=7cm]{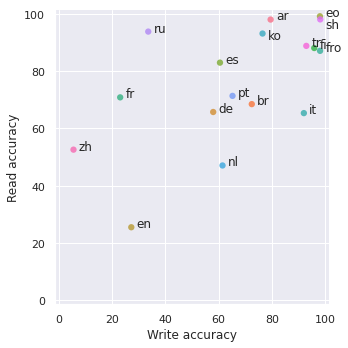}%
}{%
  \caption{Scores with $3,000$ training samples.}%
  \label{fig:figure_results_3000}
}
\end{figure}
\begin{figure}
\centering
\captionsetup{justification=centering}
\ffigbox{%
\includegraphics[width=7cm]{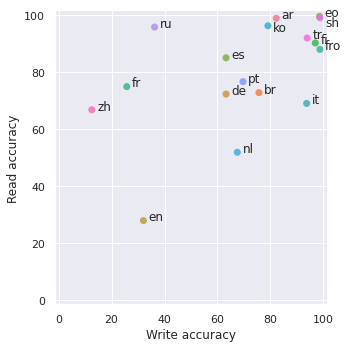}%
}{%
  \caption{Scores with $5,000$ training samples.}%
  \label{fig:figure_results_5000}
}
\end{figure}

\begin{figure}
\centering
\captionsetup{justification=centering}
\ffigbox{%
\includegraphics[width=7cm]{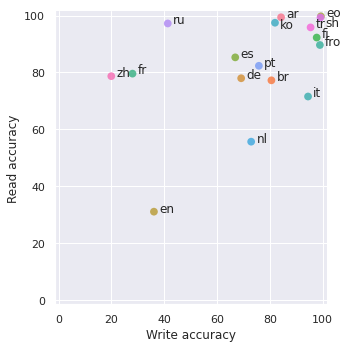}%
}{%
  \caption{Scores with $10,000$ training samples.}%
  \label{fig:figure_results_10000}
}
\end{figure}

\end{document}